\documentclass[a4paper]{article}

\usepackage{INTERSPEECH2022}

\usepackage{multirow}
\usepackage{graphicx}
\usepackage{hyperref}
\usepackage{color}
\usepackage[normalem]{ulem}

\usepackage{comment}

\title{VocaLiST: An Audio-Visual Synchronisation Model for Lips and Voices}
\name{Venkatesh S. Kadandale, Juan F. Montesinos, Gloria Haro}
\address{
  Dept. of Information and Communication Technologies, Universitat Pompeu Fabra, Spain}
\email{\{venkatesh.kadandale, juanfelipe.montesinos, gloria.haro\}@upf.edu}

\begin{document}

\maketitle
\begin{abstract}
    In this paper, we address the problem of lip-voice synchronisation in videos containing human face and voice. Our approach is based on determining if the lips motion and the voice in a video are synchronised or not, depending on their audio-visual correspondence score. We propose an audio-visual cross-modal transformer-based model that outperforms several baseline models in the audio-visual synchronisation task on the standard lip-reading speech benchmark dataset LRS2. While the existing methods focus mainly on lip synchronisation in speech videos, we also consider the special case of the singing voice. The singing voice is a more challenging use case for synchronisation due to sustained vowel sounds. We also investigate the relevance of lip synchronisation models trained on speech datasets in the context of singing voice. Finally, we use the frozen visual features learned by our lip synchronisation model in the singing voice separation task to outperform a baseline audio-visual model which was trained end-to-end.
    The demos, source code, and the pre-trained models are available on \url{https://ipcv.github.io/VocaLiST/}
    
\end{abstract}

\section{Introduction}
Online media is brimming with user-generated videos involving human voice in the form of speech and singing voice. A large number of these videos suffer from misalignment between the audio and visual streams arising due to the lack of appropriate care during video editing. As a result of this misalignment, the video viewers notice that the lip motion is not perfectly synchronised with the voice in the audio. Lip-voice synchronisation in such videos could be corrected by compensating the offset between the audio and visual modalities, which is rendering the video out-of-sync. A common method to determine this offset is by training a model to tell lip-synchronised audio-visual (AV) pairs from the non-synchronised pairs and then choose the alignment between the audio and visual signals in the non-synchronised pairs that maximise the synchronisation score. We design and train a new model that can discriminate between synchronised and unsynchronised AV pairs by learning their mutual AV correspondence in a self-supervised fashion.

A system capable of detecting lip synchronisation issues in a video has many applications. Such a system can be deployed to automatically fix the synchronisation error in online user-generated videos that suffer from this problem. It can also be used to automatically evaluate the performance of automatic AV dubbing of movies from one language to another. The features learned by the model trained to detect synchronisation have already been shown to be useful for other AV applications like visual speech recognition \cite{chung2019perfect}, cross-modal retrieval \cite{chung2020perfect}, same language dubbing \cite{shalev2020end}, dubbing \cite{prajwal2020lip}, source separation \cite{afouras2020self, owens2018audio, zhu2022visually, pan2022selective} and source localisation \cite{afouras2020self,owens2018audio}. %

We make several contributions through this paper. We propose a novel AV transformer-based model for detecting lip-voice synchronisation that estimates the extent of synchronisation between the lips motion and the voice in a given voice video. Our lip synchronisation detector outperforms state-of-the-art models on the speech benchmark dataset Lip Reading Sentences 2 (LRS2) \cite{afouras2018deep}. Besides, we also train and test our model on the Acappella \cite{montesinos} dataset which contains videos of solo singing performances. We are the first to explore the relevance of lip-sync detectors trained in speech videos to detect lip-sync in the singing voice. We use the learned visual features in the lip synchronisation detection task to outperform a singing voice separation baseline model to showcase the practical utility value of our work. The source code and the pre-trained weights of our model will be provided for reproducibility.

The rest of the paper is organised as follows. We discuss the related work in the next section. In Sec.\ 3, we delineate our network architecture. Sec.\ 4 is dedicated to the experimentation, results, and discussion. Finally, we end the paper with a conclusion section and pointers for future work.

\section{Related Work}
The seminal work \textit{Audio vision} \cite{hershey1999audio} used AV synchronisation, measured as the mutual information between non-learned audio and visual features, to localise the active speaker in a video. More recently, with the advent of deep learning techniques, different models have addressed the synchronisation of AV signals in a self-supervised way, by automatically creating positive and negative AV sync/out-of-sync pairs. While some works focus on synchronisation in general sounds \cite{owens2018audio, korbar2018cooperative, chen2021audio}, others are specialised in speech signals \cite{chung2019perfect,shalev2020end,prajwal2020lip,chung2016out,kim2021end}. 
The common trend is to extract features from each modality with an audio and a visual stream and then measure similarity/distance between the two embeddings using a sliding window to infer the offset of synchronisation. The first deep learning-based model for AV synchronisation in speech \cite{chung2016out} uses a contrastive loss with a positive and a negative pair. 
The use of $N$ negative pairs in a multi-way cross-entropy loss with softmax function \cite{chung2019perfect} further improved the performance of the same model. Unlike these works, \cite{kim2021end} directly train the model to determine the offset in the AV pairs.

Transformers have emerged as powerful deep learning architectures capable of capturing long-range dependencies in time series. Lately, transformers have been explored for several AV tasks such as source separation \cite{truong2021right, montesinos2022vovit}, source localisation \cite{lin2020audiovisual} and speech recognition \cite{ma2021end}, including synchronisation \cite{chen2021audio}. Our work in this paper is closest to Audio-Visual Synchronisation with Transformers (AVST) \cite{chen2021audio}. At the high-level, both the models share the same outline as shown in the left part of Fig.~\ref{fig:vocalist}. However, the overall architecture of our model is different with regard to the choice of audio and visual encoders and the design of the synchronisation block. The details of our model architecture are clearly outlined in Sec.\ 3.1. Besides, \cite{chen2021audio} use the InfoNCE \cite{oord2018representation} loss for optimising their model, we use binary cross entropy loss. The main focus in \cite{chen2021audio} is AV synchronisation in general audio classes, while we focus on lip-synchronisation in speech and singing voice videos only. Finally, unlike in \cite{chen2021audio}, we demonstrate a real-world application of the learned features of our synchronisation model. 

\section{Method}
\subsection{Architecture}
\begin{figure}[t]
  \centering
  \includegraphics[width=\linewidth]{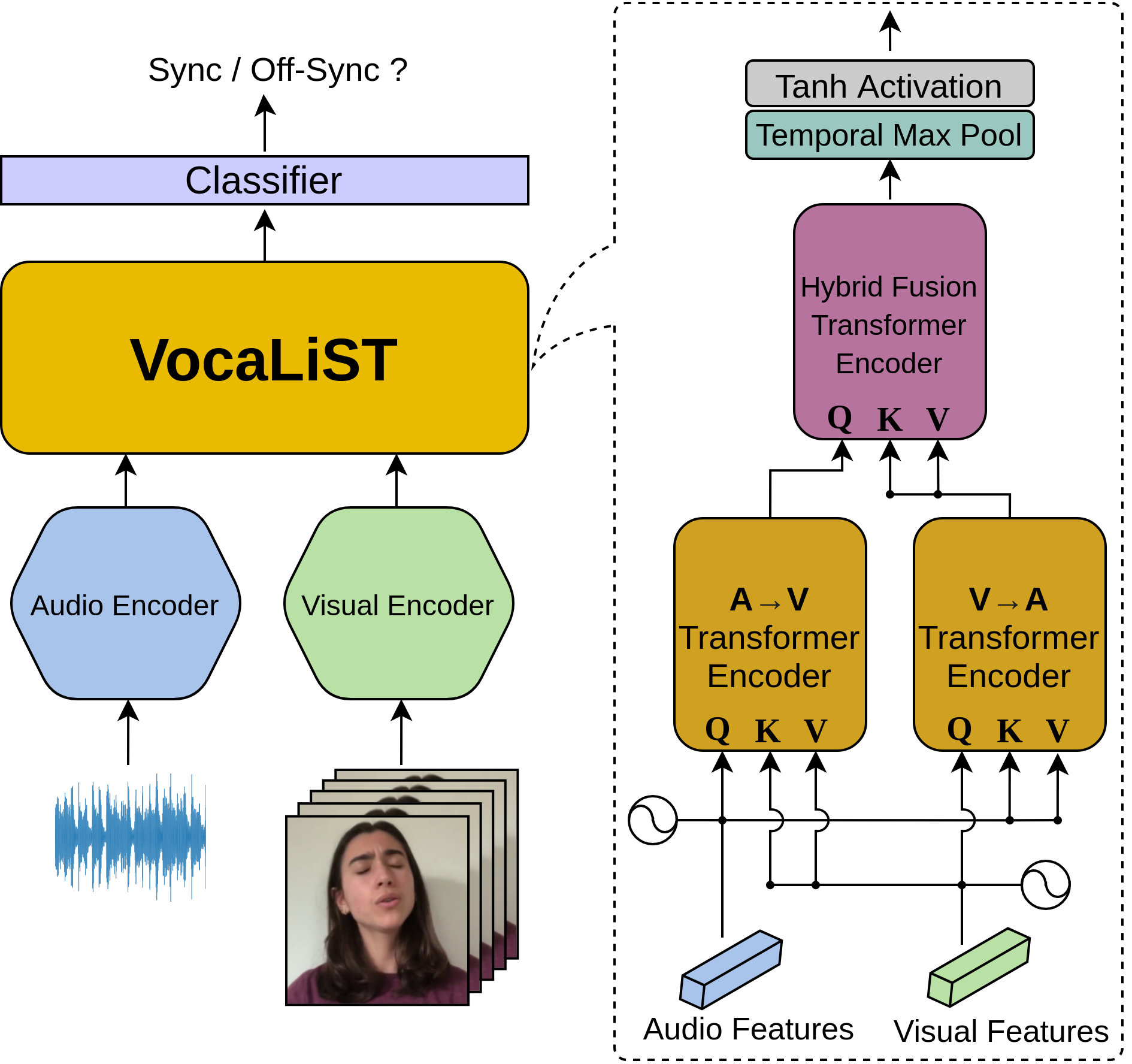}
  \caption{Architecture of our lip synchronisation detector.}
  \label{fig:vocalist}
\end{figure}

The architecture of our model is shown in Fig.\ \ref{fig:vocalist}. 
We use a transformer-based classification model which uses the audio and visual features estimated by the audio and visual encoders to classify if a given face-voice pair is synchronised or not.

\subsubsection{Audio Encoder}
\begin{figure}[t]
  \centering
  \includegraphics[width=0.8\linewidth]{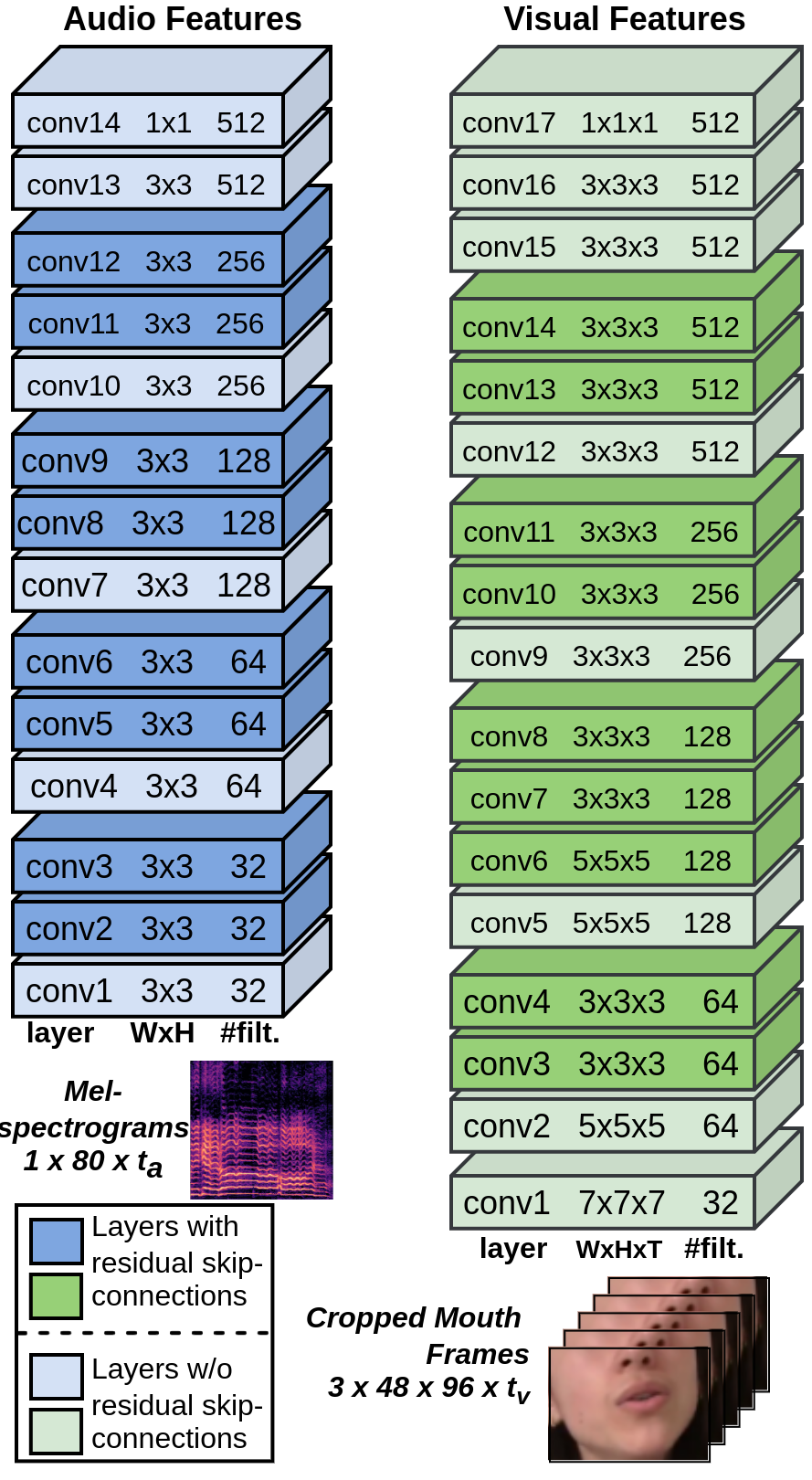}
  \caption{Our audio encoder (left) and visual encoder (right).}
  \label{fig:av_encoder}
\end{figure}

Our audio encoder (see Fig.~\ref{fig:av_encoder}) is very similar to the one used in the lip-sync expert discriminator \cite{prajwal2020lip}. The audio encoder is a stack of 2D convolutional layers with residual skip-connections that operate on the mel-spectrogram input of dimensions $1 \times 80 \times t_a$. The mel-spectrograms are obtained using 80 mel filterbanks with a hop size of 200 and window size of 800. The audios have a 16kHz sampling rate. The audio features are of the dimension $512 \times t_a$. The audio encoder conserves the temporal resolution of the input.

\subsubsection{Visual Encoder}
The visual encoder ingests a sequence of RGB images cropped around the mouth having dimensions $3 \times 48 \times 96 \times t_v$. Its architecture is inspired by the visual encoder of the lip-sync expert discriminator. Unlike in the latter, we apply 3D convolutions and conserve the temporal resolution in the feature maps. The output visual features are of  dimension $512 \times t_v$. The conservation of temporal resolution in both the audio and visual features is helpful for learning the synchronisation patterns between the two modalities spread across the temporal dimension when we feed them into the synchronisation module. The visual frames are sampled from videos of 25 fps.

\subsubsection{Synchronisation block}
We design a powerful cross-modal AV transformer that can use the AV representations learned in its cross-modal attention modules to deduce the inherent AV correspondence in a synchronised voice and lips motion pair. We refer to our transformer model as VocaLiST, the \textbf{Voca}l \textbf{Li}p \textbf{S}ync \textbf{T}ransformer. Its design is inspired by the cross-
modal transformer from \cite{tsai2019multimodal}. The cross-modal attention blocks track correlations between signals across modalities.  

The synchronisation block contains three cross-modal transformer encoders, each made up of 4 layers, 8 attention heads, and the hidden unit dimension of 512. The A$\to$V unit takes in audio features as the query and the visual features as the key and values. The roles of these audio and visual features are swapped in the V$\to$A unit. The output of the A$\to$V unit forms the query to the hybrid fusion transformer unit, while its key and values are sourced from the output of the V$\to$A unit. We max-pool the output of this hybrid fusion unit along the temporal dimension and pass it through \textit{tanh} activation. Fig.~\ref{fig:vocalist} shows the architecture of VocaLiST.

Finally, there is a fully-connected layer acting as a classifier which outputs a score indicating if the voice and lips motion is synchronised or not. The whole architecture can handle any length of audio and visual inputs.

\subsection{Training Setup}
We train our model for estimating the AV correspondence score for a given AV pair in an end-to-end manner. The positive examples correspond to the synchronised pairs in which the audio corresponds to the visual. The negative examples are obtained by introducing random temporal misalignment between the in-sync AV pairs. This allows us to follow a self-supervised training pipeline. During the training, the positive and the negative examples are sampled equally. The AV correspondence score needs to be maximised for the positive examples and minimised for the negative examples. The binary cross-entropy loss criterion is used to optimise the model parameters during the training. Unless stated otherwise, we train our model with an AV input corresponding to a sequence length of 5 visual frames (0.2s) sampled at 25 fps. We use this training setup in 4.1 and 4.2.

\section{Experiments}
Here, we investigate lip synchronisation in two different settings: speech and singing voice. We also show the usefulness of the lip synchronisation models for a practical AV application.

\subsection{Lip Synchronisation in Speech}
\textbf{Dataset.} For the task of lip synchronisation in speech videos, we consider the LRS2 \cite{afouras2018deep} dataset. The baseline methods SyncNet \cite{chung2016out}, Perfect Match (PM) \cite{chung2019perfect} and Audio-Visual Synchronisation with Transformers (AVST) \cite{chen2021audio} also use the same dataset. This allows us to directly compare our method against these baselines. As in the baseline methods, we train our model on the `pretrain' subset of LRS2 and evaluate on the `test' subset.

\noindent \textbf{Evaluation protocol.} For a fair comparison, we mimic the evaluation protocol followed by the baseline methods. Given a sequence of cropped mouth frames and the mel-spectrograms, our model is tasked with predicting the correct synchronisation in the given pair of inputs. Human observers cannot tell unsynchronised speech videos from the synchronised ones when the temporal offset is in the $\pm$1 frame range. Hence, for a synchronised pair of speech audio and lips motion inputs, the synchronisation is estimated to be correct if the predicted offset between the pair is within $\pm$ 1 frame range. As in the baseline models, we estimate this offset by finding the index of maximum AV correspondence between the set of 5-frame visual features and all the audio feature sets (each temporally matching the length of 5 visual frames) lying in the $\pm$15 frame range. 

Though we mainly train our model on AV input pairs corresponding to the length of 5 visual frames, the model can be tested on inputs of larger lengths. This is particularly useful when a set of 5 visual frames is non-informative (e.g. silence in speech). To evaluate on inputs of a larger context window, the AV correspondence score is obtained for each possible 5-frame set within the evaluation window (with a temporal stride of 1 frame) and then averaged before determining the offset. This averaging has been shown to improve the accuracy of the baseline models. On the other hand, \cite{chen2021audio} showed that training models for specific context window sizes (other than 5) leads
to better performance than the previous averaging strategy in the same context window.

\begin{table}[h]
\caption{Accuracy of lip synchronisation models in LRS2}
\label{tab:lrs2sync}
\resizebox{\linewidth}{!}{%
\begin{tabular}{|c|c|cccccc|}
\hline
\multirow{2}{*}{Models} & \multirow{2}{*}{\# params} & \multicolumn{6}{c|}{Clip Length in frames (seconds)} \\ \cline{3-8} 
& & \multicolumn{1}{c|}{5 (0.2s)} & \multicolumn{1}{c|}{7 (0.28s)} & \multicolumn{1}{c|}{9 (0.36s)} & \multicolumn{1}{c|}{11 (0.44s)} & \multicolumn{1}{c|}{13 (0.52s)} & 15 (0.6s) \\ \hline
SyncNet \cite{chung2016out} & 13.6M & \multicolumn{1}{c|}{75.8} & \multicolumn{1}{c|}{82.3} &  \multicolumn{1}{c|}{87.6} & \multicolumn{1}{c|}{91.8} & \multicolumn{1}{c|}{94.5} & 96.1 \\ 
PM \cite{chung2019perfect} & 13.6M &  \multicolumn{1}{c|}{88.1} & \multicolumn{1}{c|}{93.8} & \multicolumn{1}{c|}{96.4} & \multicolumn{1}{c|}{97.9} & \multicolumn{1}{c|}{98.7} & 99.1 \\ 
AVST \cite{chen2021audio} & 42.4M & \multicolumn{1}{c|}{92.0} & \multicolumn{1}{c|}{95.5} & \multicolumn{1}{c|}{97.7} & \multicolumn{1}{c|}{98.8} & \multicolumn{1}{c|}{99.3} & 99.6 \\ 
VocaLiST & 80.1M & \multicolumn{1}{c|}{\textbf{92.8}} & \multicolumn{1}{c|}{\textbf{96.7}} & \multicolumn{1}{c|}{\textbf{98.4}} & \multicolumn{1}{c|}{\textbf{99.3}} & \multicolumn{1}{c|}{\textbf{99.6}} & \textbf{99.8} \\ \hline
\multicolumn{4}{l}{\!\!\!(M = million)}
\end{tabular}%
}
\end{table}

\noindent \textbf{Results and discussion.} Table \ref{tab:lrs2sync} shows a direct comparison of the lip synchronisation accuracy of our model against the baseline models on the LRS2 test set. The accuracy is computed for multiple context windows ranging from sizes of 5 to 15. The accuracy improves across all the models as we increase the context window size while the rate of improvement reduces.

For each of the context window sizes, our model outperforms all the listed baseline models on the LRS2 test set. This improved performance can be attributed to the powerful cross-modal transformer blocks in our model in combination with our 3D-CNN based visual encoder. We tried reducing the complexity of our model by eliminating up to 2 cross-modal transformer blocks and also by replacing our visual encoder with the 18-layered Mixed Convolution network \cite{tran2018closer}. Both these attempts resulted in poorer performance.

\subsection{Lip Synchronisation in Singing Voice}\label{sec:sync_singing}
\textbf{Dataset.} We use Acappella \cite{montesinos}, the only publicly available AV singing voice dataset. It consists of around 46-hours of solo-singing videos spanning four language categories. We test on the unseen-unheard test subset of the dataset which contains 83 song performances evenly distributed across each of the language categories and the genders. All the videos in the Acappella dataset have been manually curated from YouTube with care taken to ensure that the samples appear synchronised to human visual observation. On the other hand, in LRS2, an automatic method was used to better synchronise the AV signal pairs.

\noindent \textbf{Models.} We consider two models for the singing voice lip synchronisation: a baseline model and our VocaLiST. Instead of selecting SyncNet \cite{chung2016out} as the baseline model, we choose the Lip Sync Expert Discriminator \cite{prajwal2020lip}. We refer to this new baseline model as SyncNet*. SyncNet* improves upon its predecessor SyncNet in the following ways: Unlike in SyncNet, i) SyncNet* operates on the RGB images, ii) the model is significantly deeper than the former with residual skip connections, and iii) the model is optimised using a cosine-similarity distance metric in combination with binary cross-entropy loss. We train both our VocaLiST and SyncNet* on the Acappella dataset's training set in an end-to-end manner.

\noindent \textbf{Evaluation protocol.} Unlike in speech videos, it is difficult for humans to notice out-of-sync in videos belonging to the general sound categories \cite{chen2021audio} when the offset between the modalities is less than 5. Singing voice also falls in such general sound category. The presence of sustained vowel sounds in singing voice, makes it difficult to notice synchronisation errors for the offsets less than 5. Thus, we consider synchronisation to be correctly estimated if the model predicts the maximum AV correspondence within the $\pm$5 frame range with respect to the ground truth. Unlike in \cite{chen2021audio}, we do not decode the videos in lower frame rate for evaluation in singing voice.
 
\begin{table}[h]
\caption{Accuracy of lip synchronisation in Acappella dataset}
\label{tab:singing_voice_perf}
\resizebox{\linewidth}{!}{%
\begin{tabular}{|c|c|c|ccccc|}
\hline
\multirow{2}{*}{Models} & \multirow{2}{*}{Var} & \multirow{2}{*}{Trained on}
 & \multicolumn{5}{c|}{Clip Length in frames (seconds)} \\ \cline{4-8} 
 & &  &  \multicolumn{1}{c|}{5 (0.2s)} & \multicolumn{1}{c|}{10 (0.4s)} & \multicolumn{1}{c|}{15 (0.6s)} & \multicolumn{1}{c|}{20 (0.8s)} & 25 (1s) \\ \hline
SyncNet* & N & Acappella & \multicolumn{1}{c|}{57.7} & \multicolumn{1}{c|}{63.9} & \multicolumn{1}{c|}{69.9} & \multicolumn{1}{c|}{75.1} & 78.7\\ 
SyncNet* & Y & Acappella & \multicolumn{1}{c|}{57.7} & \multicolumn{1}{c|}{65.9} & \multicolumn{1}{c|}{---} & \multicolumn{1}{c|}{---} & 73.6 \\ 
VocaLiST & N & LRS2 & \multicolumn{1}{c|}{56.7} & \multicolumn{1}{c|}{65.1} & \multicolumn{1}{c|}{72.2} & \multicolumn{1}{c|}{77.2} &  81.2\\
VocaLiST & N & Acappella & \multicolumn{1}{c|}{58.8} & \multicolumn{1}{c|}{65.4} & \multicolumn{1}{c|}{71.6} & \multicolumn{1}{c|}{76.5} &  80.5\\
VocaLiST & Y & Acappella & \multicolumn{1}{c|}{\textbf{58.8}} & \multicolumn{1}{c|}{\textbf{66.4}} & \multicolumn{1}{c|}{---} & \multicolumn{1}{c|}{---} & \textbf{85.2}\\ \hline
\end{tabular}%
}
\end{table}

\noindent \textbf{Results and discussion.}
We show the results in Table \ref{tab:singing_voice_perf}. Firstly, we test the models trained on the speech dataset LRS2 directly on the singing voice samples. Among the baseline models discussed in Table \ref{tab:lrs2sync}, only the SyncNet model is publicly available. The Var column indicates if the results on larger context windows are obtained by the model trained with the exact length of the context window as the ones used for testing. Several observations arise from the results. First, the synchronisation accuracy in singing voice is lower than in the case of speech, even with a larger offset tolerance, showing that singing voice synchronisation is a harder task. Also, the increase of the context window size supposes a larger improvement, compared to the speech case. The best results are achieved with a dedicated network trained for 1s-length window. It can be noticed how our model trained for speech synchronisation generalises quite well for the singing voice case with large enough context window sizes. We hypothesise that with larger contexts it is more probable to find portions of singing voice excerpt with non-sustained vowels, or consonants, producing audio-visual cues more similar to the ones found in speech. The LRS2 `pretrain' subset that we use for training spans around 195 hours, while the training set of Acappella totals up to less than 37 hours of videos. Hence, compared to LRS2, Acappella is a small dataset and, therefore, the models trained on LRS2 tend to perform better in Acappella than the other way round.


\subsection{Singing Voice Separation}
We would like to demonstrate the effectiveness of the features learned by the visual encoder of our synchronisation network by using them in the singing voice separation task.

\noindent \textbf{Dataset.} We use the Acappella dataset \cite{montesinos}, as in Sec.~\ref{sec:sync_singing}.

\noindent \textbf{Training.} We use the same training pipeline that is used for training the model Y-Net-mr in  \cite{montesinos}. Y-Net-mr is a U-Net \cite{ronneberger2015u} conditioned by visual features extracted from cropped mouth frames using an 18-layer mixed convolution network \cite{tran2018closer}. Y-Net-mr estimates the complex masks corresponding to a voice present in an audio mixture when given with a spectrogram of an audio mixture and the temporal sequence of the cropped mouth frames corresponding to the target voice as input. During training, half of the input audio mixtures contain one voice mixed with a musical accompaniment, while the other half contains an additional voice besides the accompaniment.  

\noindent \textbf{Models.} The main baseline model here is the Y-Net-mr from  \cite{montesinos} which has been trained end-to-end for singing voice separation. It is a state-of-the-art audio-visual singing voice separation model among the models that operate directly on the cropped mouth frames. To investigate the contribution of the features learned by our lip synchronisation model, we propose Y-Net-mr-V. Y-Net-mr-V is nothing but a Y-Net-mr with its visual encoder replaced with that of our lip synchronisation model. In one setting, we train the Y-Net-mr-V for singing voice separation by loading the pre-trained weights of the visual encoder which was trained as a part of VocaLiST for singing voice lip synchronisation and keep them frozen throughout the training. In another setting, we train the entire model Y-Net-mr-V in an end-to-end manner without using the pre-trained weights learned from the lip synchronisation task. Finally, we also consider Y-Net-mr-S*, which is a Y-Net-mr with its visual encoder replaced by that of the SyncNet* \cite{prajwal2020lip}. For Y-Net-mr-S*, we only consider the setting where the pre-trained visual encoder weights corresponding to the singing voice lip synchronisation task are loaded from the SyncNet* and frozen as we train the rest of Y-Net-mr-S* for singing voice separation.

\noindent \textbf{Evaluation.} We evaluate the singing voice separation performance by computing the source separation metrics \cite{vincent2006performance} Source-to-Distortion Ratio (SDR) and  Source-to-Interference Ratio (SIR) on the estimated target voices in the unseen-unheard test subset of the Acappella dataset in the two-lead voice setup. In the two-lead voice setup, the music mixtures are artificially generated by randomly mixing two solo singing voices from the Acappella dataset in addition to another accompaniment audio. The higher these metrics are, the better the performance.

\begin{table}[h]
\caption{Performance metrics for Singing Voice Separation}
\label{tab:source_sep}
\resizebox{\linewidth}{!}{%
\begin{tabular}{|c|c|c|c|}
\hline
\multicolumn{1}{|c|}{\multirow{2}{*}{Architecture}} & \multicolumn{1}{c|}{\multirow{2}{*}{Method}} & \multicolumn{2}{c|}{Source Separation Metrics} \\ \cline{3-4} 
  &  & \multicolumn{1}{c|}{\hspace*{3.3mm} SDR \hspace*{3.3mm}} & \multicolumn{1}{c|}{SIR}\\ \hline
Y-Net-mr \cite{montesinos} & E2E & \multicolumn{1}{c|}{5.03} & \multicolumn{1}{c|}{15.80}\\
Y-Net-mr-V & E2E & \multicolumn{1}{c|}{1.14} & \multicolumn{1}{c|}{11.72}\\
Y-Net-mr-S* & PT - SyncNet* & \multicolumn{1}{c|}{5.44} & \multicolumn{1}{c|}{16.17}\\
Y-Net-mr-V & PT - VocaLiST & \multicolumn{1}{c|}{\textbf{6.32}} & \multicolumn{1}{c|}{\textbf{17.08}}\\ \hline
%
\end{tabular}%
}
\end{table}

\noindent \textbf{Results and Discussion.}
Table \ref{tab:source_sep} shows the performance of the models in singing voice separation. Our model Y-Net-mr-V outperforms Y-Net-mr when we use the pre-trained weights from the lip synchronisation task for the visual encoder. It is challenging to train a source separation network that operates directly with video frames when the dataset is small, which is the case of Acappella. There is a tendency of over-fitting in such small datasets. Under such circumstances, knowledge transfer from the audio-visual synchronisation could guide the source separation despite the dataset size limitations. To further highlight the importance of this knowledge transfer, we also train the Y-Net-mr-V end to end without using the pre-trained synchronisation visual features. In this case, the model does not generalise well to the test-unseen subset despite having 38M trainable parameters solely in the visual encoder. Note that even Y-Net-mr-S* outperforms Y-Net-mr here. But since SyncNet* didn't perform as good as the VocaLisT in lip synchronisation task (see Table \ref{tab:singing_voice_perf}), as we expected, Y-Net-mr-S* did not outperform Y-Net-mr-V trained with knowledge transfer. 

\section{Conclusions}
This paper presents VocaLiST, a transformer-based model for voice-lip  synchronisation. The model has been analysed both in speech and singing voice, producing state-of-the-art results. We have shown that it learns powerful visual features that are useful for solving the problem of singing voice separation in a mixture with more than one voice. It could perform even better for larger input contexts if we have dedicated models trained for each specific length of the input,  as shown in Table \ref{tab:singing_voice_perf} and also in \cite{chen2021audio}. 

\section{Acknowledgements}

We would like to thank Honglie Chen (VGG, Oxford) and Soo-Whan Chung (Naver Corporation) for the insightful discussions. 
We acknowledge support by MICINN/FEDER UE project PID2021-127643NB-I00; H2020-MSCA-RISE-2017 project  777826 NoMADS. 
 V.\ S.\ K.\ has received support through “la Caixa” Foundation (ID 100010434), fellowship code: LCF/BQ/DI18/11660064 
and the Marie SkłodowskaCurie grant agreement No.\ 713673. J.\ F.\ M.\ acknowledges support by FPI scholarship PRE2018-083920.
We gratefully acknowledge NVIDIA Corporation for the donation of GPUs used for the experiments.

\bibliographystyle{IEEEtran}

\bibliography{mybib}

\begin{thebibliography}{10}
\providecommand{\url}[1]{#1}
\csname url@samestyle\endcsname
\providecommand{\newblock}{\relax}
\providecommand{\bibinfo}[2]{#2}
\providecommand{\BIBentrySTDinterwordspacing}{\spaceskip=0pt\relax}
\providecommand{\BIBentryALTinterwordstretchfactor}{4}
\providecommand{\BIBentryALTinterwordspacing}{\spaceskip=\fontdimen2\font plus
\BIBentryALTinterwordstretchfactor\fontdimen3\font minus
  \fontdimen4\font\relax}
\providecommand{\BIBforeignlanguage}[2]{{%
\expandafter\ifx\csname l@#1\endcsname\relax
\typeout{** WARNING: IEEEtran.bst: No hyphenation pattern has been}%
\typeout{** loaded for the language `#1'. Using the pattern for}%
\typeout{** the default language instead.}%
\else
\language=\csname l@#1\endcsname
\fi
#2}}
\providecommand{\BIBdecl}{\relax}
\BIBdecl

\bibitem{chung2019perfect}
S.-W. Chung, J.~S. Chung, and H.-G. Kang, ``Perfect match: Improved cross-modal
  embeddings for audio-visual synchronisation,'' in \emph{ICASSP 2019-2019 IEEE
  International Conference on Acoustics, Speech and Signal Processing
  (ICASSP)}.\hskip 1em plus 0.5em minus 0.4em\relax IEEE, 2019, pp. 3965--3969.

\bibitem{chung2020perfect}
S.~W. Chung, J.~S. Chung, and H.-G. Kang, ``Perfect match: Self-supervised
  embeddings for cross-modal retrieval,'' \emph{IEEE Journal of Selected Topics
  in Signal Processing}, vol.~14, no.~3, pp. 568--576, 2020.

\bibitem{shalev2020end}
Y.~Shalev and L.~Wolf, ``End to end lip synchronization with a temporal
  autoencoder,'' in \emph{Proceedings of the IEEE/CVF Winter Conference on
  Applications of Computer Vision}, 2020, pp. 341--350.

\bibitem{prajwal2020lip}
K.~Prajwal, R.~Mukhopadhyay, V.~P. Namboodiri, and C.~Jawahar, ``A lip sync
  expert is all you need for speech to lip generation in the wild,'' in
  \emph{Proceedings of the 28th ACM International Conference on Multimedia},
  2020, pp. 484--492.

\bibitem{afouras2020self}
T.~Afouras, A.~Owens, J.~S. Chung, and A.~Zisserman, ``Self-supervised learning
  of audio-visual objects from video,'' in \emph{European Conference on
  Computer Vision}.\hskip 1em plus 0.5em minus 0.4em\relax Springer, 2020, pp.
  208--224.

\bibitem{owens2018audio}
A.~Owens and A.~A. Efros, ``Audio-visual scene analysis with self-supervised
  multisensory features,'' in \emph{Proceedings of the European Conference on
  Computer Vision (ECCV)}, 2018, pp. 631--648.

\bibitem{zhu2022visually}
L.~Zhu and E.~Rahtu, ``Visually guided sound source separation and localization
  using self-supervised motion representations,'' in \emph{Proceedings of the
  IEEE/CVF Winter Conference on Applications of Computer Vision}, 2022, pp.
  1289--1299.

\bibitem{pan2022selective}
Z.~Pan, R.~Tao, C.~Xu, and H.~Li, ``Selective listening by synchronizing speech
  with lips,'' \emph{IEEE/ACM Transactions on Audio, Speech, and Language
  Processing}, 2022.

\bibitem{afouras2018deep}
T.~Afouras, J.~S. Chung, A.~Senior, O.~Vinyals, and A.~Zisserman, ``Deep
  audio-visual speech recognition,'' \emph{IEEE transactions on pattern
  analysis and machine intelligence}, 2018.

\bibitem{montesinos}
J.~F. Montesinos, V.~S. Kadandale, and G.~Haro, ``A cappella: Audio-visual
  singing voice separation,'' in \emph{32nd British Machine Vision Conference,
  BMVC}, 2021.

\bibitem{hershey1999audio}
J.~Hershey and J.~Movellan, ``Audio vision: Using audio-visual synchrony to
  locate sounds,'' \emph{Advances in neural information processing systems},
  vol.~12, 1999.

\bibitem{korbar2018cooperative}
B.~Korbar, D.~Tran, and L.~Torresani, ``Cooperative learning of audio and video
  models from self-supervised synchronization,'' \emph{Advances in Neural
  Information Processing Systems}, vol.~31, 2018.

\bibitem{chen2021audio}
H.~Chen, W.~Xie, T.~Afouras, A.~Nagrani, A.~Vedaldi, and A.~Zisserman,
  ``Audio-visual synchronisation in the wild,'' \emph{arXiv preprint
  arXiv:2112.04432}, 2021.

\bibitem{chung2016out}
J.~S. Chung and A.~Zisserman, ``Out of time: automated lip sync in the wild,''
  in \emph{Asian conference on computer vision}.\hskip 1em plus 0.5em minus
  0.4em\relax Springer, 2016, pp. 251--263.

\bibitem{kim2021end}
Y.~J. Kim, H.~S. Heo, S.-W. Chung, and B.-J. Lee, ``End-to-end lip
  synchronisation based on pattern classification,'' in \emph{2021 IEEE Spoken
  Language Technology Workshop (SLT)}.\hskip 1em plus 0.5em minus 0.4em\relax
  IEEE, 2021, pp. 598--605.

\bibitem{truong2021right}
T.-D. Truong, C.~N. Duong, H.~A. Pham, B.~Raj, N.~Le, K.~Luu \emph{et~al.},
  ``The right to talk: An audio-visual transformer approach,'' in
  \emph{Proceedings of the IEEE/CVF International Conference on Computer
  Vision}, 2021, pp. 1105--1114.

\bibitem{montesinos2022vovit}
J.~F. Montesinos, V.~S. Kadandale, and G.~Haro, ``Vovit: Low latency
  graph-based audio-visual voice separation transformer,'' \emph{arXiv preprint
  arXiv:2203.04099}, 2022.

\bibitem{lin2020audiovisual}
Y.-B. Lin and Y.-C.~F. Wang, ``Audiovisual transformer with instance attention
  for audio-visual event localization,'' in \emph{Proceedings of the Asian
  Conference on Computer Vision}, 2020.

\bibitem{ma2021end}
P.~Ma, S.~Petridis, and M.~Pantic, ``End-to-end audio-visual speech recognition
  with conformers,'' in \emph{ICASSP 2021-2021 IEEE International Conference on
  Acoustics, Speech and Signal Processing (ICASSP)}.\hskip 1em plus 0.5em minus
  0.4em\relax IEEE, 2021, pp. 7613--7617.

\bibitem{oord2018representation}
A.~v.~d. Oord, Y.~Li, and O.~Vinyals, ``Representation learning with
  contrastive predictive coding,'' \emph{arXiv preprint arXiv:1807.03748},
  2018.

\bibitem{tsai2019multimodal}
Y.-H.~H. Tsai, S.~Bai, P.~P. Liang, J.~Z. Kolter, L.-P. Morency, and
  R.~Salakhutdinov, ``Multimodal transformer for unaligned multimodal language
  sequences,'' in \emph{Proceedings of the conference. Association for
  Computational Linguistics. Meeting}, vol. 2019.\hskip 1em plus 0.5em minus
  0.4em\relax NIH Public Access, 2019, p. 6558.

\bibitem{tran2018closer}
D.~Tran, H.~Wang, L.~Torresani, J.~Ray, Y.~LeCun, and M.~Paluri, ``A closer
  look at spatiotemporal convolutions for action recognition,'' in
  \emph{Proceedings of the IEEE conference on Computer Vision and Pattern
  Recognition}, 2018, pp. 6450--6459.

\bibitem{ronneberger2015u}
O.~Ronneberger, P.~Fischer, and T.~Brox, ``U-net: Convolutional networks for
  biomedical image segmentation,'' in \emph{International Conference on Medical
  image computing and computer-assisted intervention}.\hskip 1em plus 0.5em
  minus 0.4em\relax Springer, 2015, pp. 234--241.

\bibitem{vincent2006performance}
E.~Vincent, R.~Gribonval, and C.~F{\'e}votte, ``Performance measurement in
  blind audio source separation,'' \emph{IEEE trans. on audio, speech, and
  language process.}, vol.~14, no.~4, pp. 1462--1469, 2006.

\end{thebibliography}

\end{document}